\documentclass[runningheads]{llncs}

\usepackage{eccv}

\usepackage{eccvabbrv}

\usepackage{graphicx}
\usepackage{booktabs}
\usepackage{amsmath}
\usepackage{amssymb}
\usepackage{bm}
\usepackage{algorithm}
\usepackage{algorithmic}
\usepackage{subcaption}
\usepackage{pifont}
\usepackage{threeparttable}
\usepackage[accsupp]{axessibility}

\usepackage{hyperref}

\usepackage{orcidlink}

\newcommand{\cmark}{\ding{51}}
\newcommand{\xmark}{\ding{55}}

\newlength{\imgw}
\usepackage{colortbl}
\begin{document}

\title{MTVDiff: Multimodal Conditional Latent Diffusion for Enhanced Thermal-to-Visible Face Translation}

\titlerunning{MTVDiff: Multimodal Conditional Latent Diffusion}

\author{Zhiyuan Xia\inst{1,2}\orcidlink{0009-0008-5123-8541} \and
Haojie Li\inst{3}\orcidlink{0009-0006-4141-1297} \and
Jingyu Lin\inst{3}\orcidlink{0000-0003-1897-4148} \and
Yiguo Qiao\inst{1,2}\orcidlink{0000-0003-4586-3152}\thanks{Corresponding author.} \and
Cunjian Chen\inst{3}\orcidlink{0000-0002-2926-9762}}

\authorrunning{Z.~Xia et al.}

\institute{School of Computer Science and Engineering, Southeast University, Nanjing, China\\
\email{220234779@seu.edu.cn}, \email{yqiao@seu.edu.cn} \and
Key Laboratory of New Generation Artificial Intelligence Technology and Its Interdisciplinary Applications (Southeast University), Ministry of Education, China \and
Monash University, Australia\\
\email{ext-haojieli@monash.edu}, \email{jingyu.lin@monash.edu}, \email{cunjian.chen@monash.edu}}

\maketitle

\begin{abstract}
Thermal-to-visible face translation presents fundamental challenges including geometric discontinuities, semantic attribute mismatches, and identity degradation. We propose MTVDiff, a novel multimodal latent diffusion framework that synergistically integrates depth and textual information to address these limitations while preserving identity characteristics. The MTVDiff framework presents three core technical contributions: (1) a Dual-Branch Cross-Attention Fusion (DBCAF) module for multi-scale thermal-depth feature extraction and fusion; (2) a Gated Text-to-Visual Feature Alignment mechanism for semantically-guided generation; and (3) Spatial Feature Transformations (SFT) for adaptive multimodal prior integration. Extensive experiments on the MCXFace and SpeakingFaces datasets demonstrate that our multimodal approach significantly outperforms existing GAN-based and diffusion-based approaches across multiple metrics, achieving substantial improvements in both image quality and face verification performance, with FID reductions of up to 48.3\% and Rank-1 accuracy improvements of up to 8.9\%. Our work provides a robust solution for face recognition systems operating under varying illumination conditions and advances the state-of-the-art in cross-spectral facial image translation through effective multimodal integration.
\end{abstract}

\section{Introduction}
\label{sec:intro}

Cross-modal image translation has emerged as a pivotal research area in computer vision, driven by the need to bridge disparate imaging modalities such as thermal infrared and visible light~\cite{AngheloneCRD25}. This technology has found critical applications in security and surveillance systems, where thermal-to-visible (T2V) face translation enables identity verification in challenging nighttime and low-light environments. While GANs~\cite{axialgan,AngheloneCFRD21} have achieved impressive results, DDPMs~\cite{ddpm} offer advantages in image quality, diversity, and training stability. However, transitioning these frameworks to cross-spectral scenarios—particularly thermal-to-visible face synthesis—remains challenging. T2V-DDPM~\cite{T2V-ddpm} laid the groundwork by demonstrating the feasibility and promise of guided DDPMs for thermal-to-visible translation. DiffTV~\cite{difftv1} built upon this by moving to the more efficient latent diffusion model (LDM)~\cite{ldm} and specifically tackling the crucial problem of identity preservation through sophisticated feature alignment and multi-stage conditioning.

Adding multimodal conditions like text and depth maps to thermal-to-visible (T2V) face translation offers significant advantages not explored in T2V-DDPM or DiffTV. Text provides direct semantic control, allowing specification of attributes such as gender or age, enhancing controllability and identity preservation. Depth information offers crucial 3D structural understanding, helping generate more geometrically accurate visible faces with improved robustness to pose and lighting variations. Such multimodal data can be readily acquired using RGB-D and thermal camera systems deployed in surveillance applications.

\begin{figure}[!htbp]
  \centering
  \includegraphics[width=\textwidth]{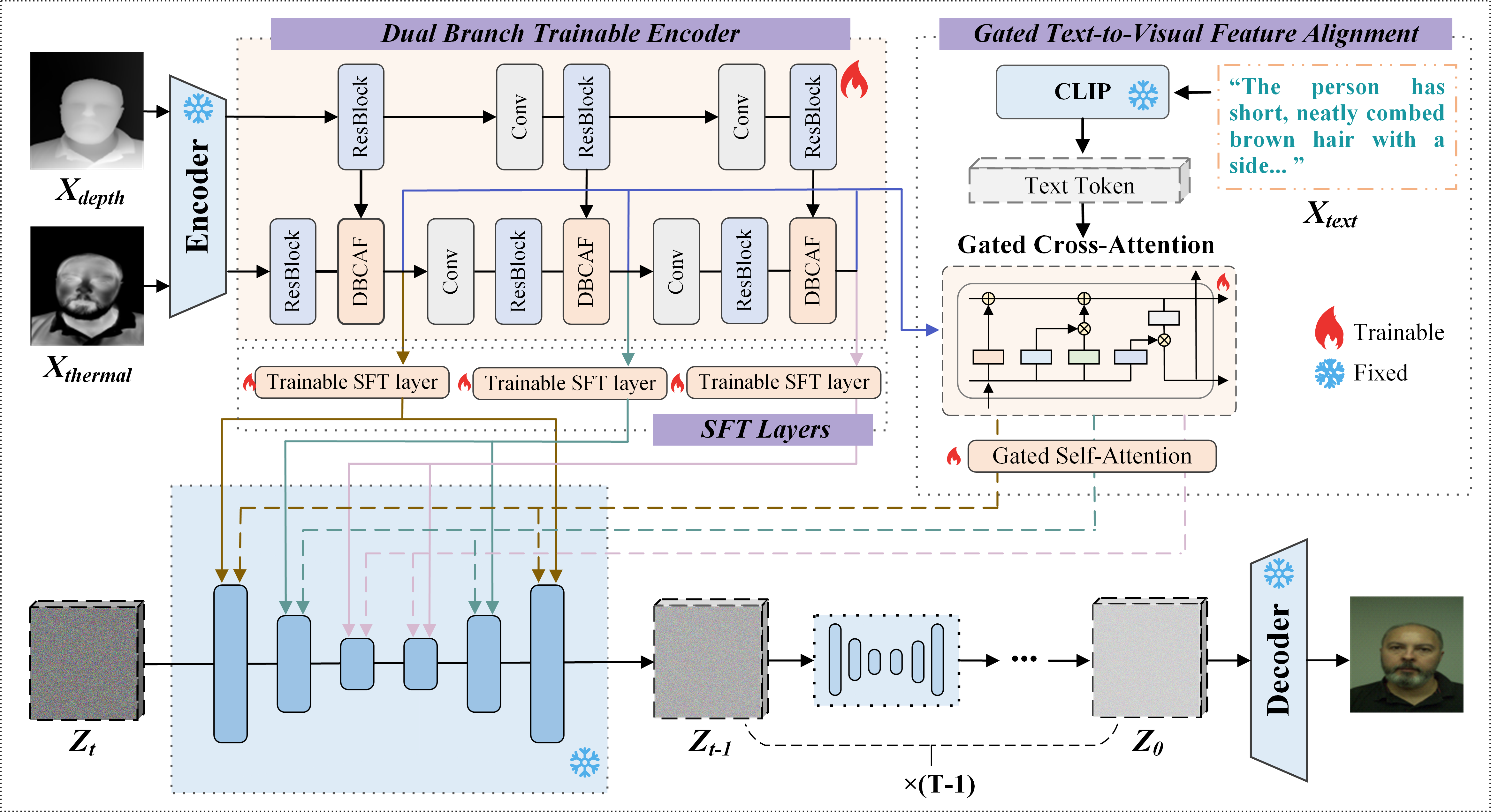}
  \caption{The overall framework of the proposed multi-modal conditional latent diffusion (MTVDiff).}
  \label{fig:overview}
\end{figure}

However, fundamental challenges remain in solving multimodal thermal-to-visible translation: (1) fixed fusion protocols inadequately resolve geometric misalignments between thermal and depth modalities; (2) conventional text prompts fail to capture intricate semantic attributes in thermal imagery; and (3) entangled feature conditioning results in illumination artifacts that compromise identity preservation. To address these challenges, we propose MTVDiff (see \cref{fig:overview}), a novel multimodal latent diffusion framework that integrates thermal images, depth information, and textual descriptions through a carefully designed pipeline. Our framework processes thermal and depth inputs through parallel encoders, dynamically fuses features using our Dual-Branch Cross-Attention Fusion (DBCAF) module, and integrates textual descriptions via a Gated Text-to-Visual Feature Alignment mechanism for semantic guidance. During diffusion, Spatial Feature Transformation injects multimodal fusion features into UNet residual blocks, preserving structural integrity while enhancing photometric details.

Our contributions are summarized as follows:
\begin{itemize}
    \item \textbf{Dual-Branch Cross-Attention Fusion:} A DBCAF module that dynamically integrates thermal and depth features through multi-level interactions and attention-adaptive weight modulation, ensuring precise geometric cross-modal alignment.
    \item \textbf{Semantically-Guided Generation:} A Gated Text-to-Visual Feature Alignment mechanism that selectively incorporates textual semantic information while preserving essential layout structures.
    \item \textbf{Adaptive Multimodal Integration:} A Spatial Feature Transformation module that injects multi-scale fusion features into diffusion residual blocks, improving identity preservation under varying illumination conditions.
\end{itemize}

\section{Related Work}
\label{sec:related}

\subsection{Image-to-Image Translation}

\noindent\textbf{GAN-based Approaches.}
Early image-to-image (I2I) translation methods predominantly employed Generative Adversarial Networks (GANs)~\cite{pix2pix,cyclegan}. Specialized architectures like Axial-GAN~\cite{axialgan} or pix2pix-zero~\cite{pix2pix-zero} integrated attention mechanisms or distillation for cross-modal translation tasks. However, GAN-based methods remain prone to training instabilities and mode collapse~\cite{mode_collapse}, facing significant challenges in bridging substantial domain gaps, particularly in thermal-to-visible translation.

\noindent\textbf{Diffusion-based Approaches.}
Recent advances in denoising diffusion probabilistic models (DDPMs)~\cite{LiX0L23,xia2024dmt,DiffI2I} offer a promising alternative for I2I translation. Latent diffusion models (LDMs)~\cite{ldm} revolutionized the field by operating in compressed latent spaces, where ControlNet~\cite{controlnet} and T2I-Adapter~\cite{T2i-adapter} were proposed to modulate the conditions. Despite these advances, application to cross-spectral translation remains underexplored, with methods like DiffV2IR~\cite{diffv2ir} often struggling to preserve identity-critical features across imaging modalities~\cite{lin2024pair,zhou2025identitystory}. Uni-ControlNet~\cite{zhao2023unicontrolnet} offers a unified multi-condition framework that supports simultaneous injection of multiple control signals into a single ControlNet, providing a natural baseline for multimodal conditional synthesis~\cite{song2025scenedecorator}; however, it was not designed for cross-spectral translation and lacks the identity-preserving mechanisms required for thermal-to-visible face synthesis.
\subsection{Thermal-to-Visible Face Translation}

Cross-spectral translation, particularly thermal-to-visible face synthesis, has evolved significantly in recent years. T2V-DDPM~\cite{T2V-ddpm} pioneered the application of diffusion models' iterative denoising to this task. Building on this, DiffTV~\cite{difftv1} advanced the field by introducing the first LDM designed for thermal-to-visible translation, employing heterogeneous feature alignment and dual-stage conditioning. Earlier methods like Axial-GAN~\cite{axialgan} addressed geometric misalignment through axial-attention layers, while DiffV2IR~\cite{diffv2ir} incorporated vision-language understanding for semantic-aware translation. However, high-fidelity visible face reconstruction from thermal data remains challenging due to sparse geometric features, pose variations, and missing chromatic information—issues that single-modal approaches struggle to fully address.

\section{Method}
\label{sec:method}

As shown in \cref{fig:overview}, the proposed MTVDiff integrates multimodal inputs through a conditional latent diffusion pipeline. Our framework processes three complementary modalities: thermal images as the primary input, depth maps for 3D structural guidance, and textual descriptions for semantic attribute control.

\subsection{Preliminary}

\noindent\textbf{Denoising Diffusion Probabilistic Models.}
Our approach utilizes the DDPM~\cite{ddpm}, which learns data distributions through iterative denoising. The forward process gradually adds Gaussian noise to data $\bm{x}_0$ until it becomes pure noise $\bm{x}_T$, while the reverse process learns to denoise through a neural network $\bm{\epsilon}_\theta$ optimized via:
\begin{equation}
\mathcal{L} = \mathbb{E}_{\bm{x}_0, t, \bm{\epsilon}}\left\lVert \bm{\epsilon} - \bm{\epsilon}_\theta(\bm{x}_t, t) \right\rVert_2^2,
\end{equation}
where $\bm{x}_0$ denotes the original clean data, $\bm{x}_t$ represents the noisy data at timestep $t$, and $\bm{\epsilon} \sim \mathcal{N}(0, \bm{I})$ is the Gaussian noise.

\noindent\textbf{Latent Diffusion Models.}
Our framework employs LDMs~\cite{ldm} that perform diffusion in compressed latent space. An encoder $\mathcal{E}$ maps input image $\bm{x}$ to latent representation $\bm{z} = \mathcal{E}(\bm{x})$, and the diffusion process optimizes:
\begin{equation}
\mathcal{L} = \mathbb{E}_{\bm{z}_0, t, \bm{\epsilon} \sim \mathcal{N}(0, \bm{I})}\left[\left\lVert \bm{\epsilon} - \bm{\epsilon}_\theta(\bm{z}_t, t, \bm{c}) \right\rVert_2^2\right],
\end{equation}
where $\bm{c}$ denotes conditional generation guidance. After training, a decoder $\mathcal{D}$ transforms the denoised latent representation back to pixel space: $\hat{\bm{x}} = \mathcal{D}(\bm{z}_0)$.

\subsection{Dual-Branch Cross-Attention Fusion}

\begin{figure}[!htbp]
  \centering
  \includegraphics[width=0.95\textwidth]{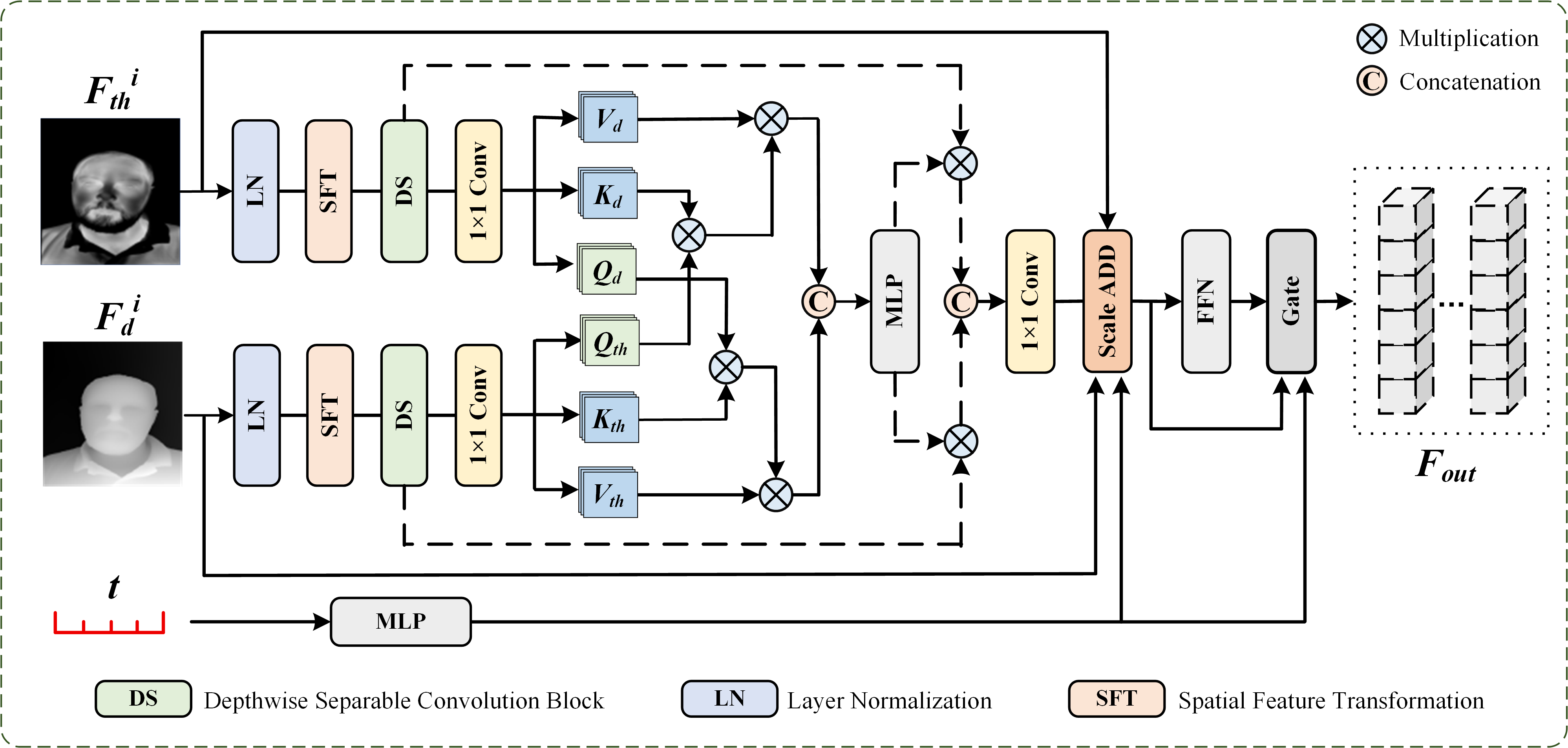}
  \caption{Illustration of the Dual-Branch Cross-Attention Fusion (DBCAF) module.}
  \label{fig:DBCAF}
\end{figure}

To address geometric boundary blurring caused by low-resolution thermal imaging, noise in sparse depth features, and rigid fusion in nighttime conditions, we propose a Dual-Branch Cross-Attention Fusion (DBCAF) module inspired by prior work~\cite{3diffusion}. As illustrated in \cref{fig:DBCAF}, operating exclusively during the downsampling phase, the DBCAF module receives multi-scale features extracted by dual ResNet-18 encoders from depth and thermal facial inputs. Each encoder processes its respective modality through a series of ResBlocks:
\begin{equation}
F_{enc} = \text{ResBlock}(x_{input}).
\end{equation}

To dynamically generate guidance information for different time steps, we feed $t$ through a multilayer perceptron (MLP) to learn the weight parameters:
\begin{equation}
s_i, b_i, g = \text{MLP}(t),
\end{equation}
where $s_i, b_i \in \mathbb{R}^d$ are scale and bias modulation signals for $i = 1, 2, 3$, and $g \in \mathbb{R}^d$ is the gating modulation signal.

We apply Spatial Feature Transformation (SFT) following layer normalization, where temporal embeddings dynamically modulate features from both branches:
\begin{equation}
F_{i} = \text{SFT}(\text{LN}(F_{d}^{i}), s_i, b_i) = s_i \odot (1 + \text{LN}(F_{d}^{i})) + b_i,
\end{equation}
where $F_{i}$ denotes the transformed output feature, and $\odot$ stands for element-wise multiplication. To model global feature interactions across modalities, we employ a cross-attention mechanism:
\begin{equation}
Z_a = \text{softmax}\left(\frac{Q_b K_a^T}{\sqrt{\hat{C}}}\right)V_a,
\end{equation}
where $Q_b \in \mathbb{R}^{N \times C'}$ denotes queries from modality $b$, while $K_a, V_a \in \mathbb{R}^{N \times C'}$ are key and value from modality $a$.

The cross-attention outputs are concatenated into $F_c$, followed by an MLP and softmax to produce channel-wise weights $w_a, w_b \in \mathbb{R}^{C\times H\times W}$:
\begin{equation}
[w_a, w_b] = \text{softmax}(\text{MLP}(F_c)).
\end{equation}
The aggregated feature map $M_l$ at the $l$-th layer is computed as:
\begin{equation}
M_l = w_a \odot F_l^a + w_b \odot F_l^b.
\end{equation}
The final output integrates the modalities through adaptive balancing:
\begin{equation}
    F_{\text{out}} = {\mathcal{C}_{1 \times 1}(x)} + {F_{\text{t}} \cdot s_3} + {F_{\text{d}} \cdot b_3},
\end{equation}
followed by a Feedforward Neural Network (FFN) and gating operation:
\begin{equation}
F_{out} = \text{FFN}(x) + g \odot x.
\end{equation}

\subsection{Gated Text-to-Visual Feature Alignment}

To leverage semantic information from textual descriptions, we encode textual descriptions into embedding tensors via CLIP's text encoder. Although standard cross-attention mechanisms in Stable Diffusion effectively incorporate text conditions, we identify limitations in the interaction between visual layout conditions and textual descriptors.

Building upon~\cite{diffx}, we insert a gated cross-attention (CA) layer followed by a Feed-Forward Network (FFN) to enhance inter-modal interaction:
\begin{equation}
\text{CA}(H, c^*) = \text{softmax}\left(\frac{W_q(H) \cdot W_k(c^*)^T}{\sqrt{d_k}}\right) \cdot W_v(c^*),
\end{equation}
where $H$ is the fused multimodal features from the DBCAF module, $c^*$ the text embeddings, and $W_q$, $W_k$, $W_v$ are learnable projection matrices. To regulate cross-modal information influence, we introduce learnable gating parameters:
\begin{align}
H' &= H + \lambda \cdot \tanh(\gamma_1) \cdot \text{CA}(H, c^*), \\
H^* &= H' + \lambda \cdot \tanh(\gamma_2) \cdot \text{FF}(H'),
\end{align}
where $\lambda$ balances feature quality and controllability, and $\gamma_1$, $\gamma_2$ are adaptive learnable scalars.

Similarly, we employ gated self-attention within the UNet architecture to integrate multimodal features:
\begin{align}
F' &= F + \lambda \cdot \tanh(\gamma_3) \cdot \text{SelfAttn}([F, \text{Proj}(H^*)]), \\
F^* &= F' + \lambda \cdot \tanh(\gamma_4) \cdot \text{MLP}(F'),
\end{align}
where $F$ denotes UNet feature representations, and $\gamma_3$, $\gamma_4$ are additional gating parameters.

\subsection{Spatial Feature Transformations}

Similar to StableSR~\cite{stablesr}, our framework introduces minimal architectural modifications to the original Stable Diffusion model. We maintain frozen weights in Stable Diffusion while exclusively training the encoder and SFT layers, thereby preserving its pretrained knowledge while enabling structural information integration. The SFT modulates Stable Diffusion's residual blocks:
\begin{equation}
\begin{aligned}
\gamma_l, \beta_l &= \text{Conv}_{1\times1}\left(F_{\text{fusion}}^{(l)}\right), \\
h'_l &= \text{GroupNorm}(h_l) \odot (1 + \gamma_l) + \beta_l,
\end{aligned}
\end{equation}
where $F_{\text{fusion}}^{(l)}$ denotes the multi-scale fusion features at layer $l$, $\gamma_l$ and $\beta_l$ are the scaling factor and bias term, and $h_{l}$ represents the original feature map in the $l$-th residual block. This approach ensures seamless multi-scale thermal-depth fusion while preserving Stable Diffusion's generative priors.

\section{Experiments}
\label{sec:experiments}

\subsection{Datasets and Evaluation Metrics}

\noindent\textbf{Datasets.}
We evaluate on two benchmarks. \textbf{MCXFace}~\cite{mcxface} contains 6,120 thermal-visible image pairs (256$\times$256) from 51 subjects captured across three sessions under varied illumination, partitioned into 41 training subjects (4,920 pairs) and 10 test subjects (1,200 pairs). \textbf{SpeakingFaces}~\cite{SpeakingFace} is a large-scale multimodal dataset comprising synchronized thermal and visual streams from 142 subjects. After quality filtering, we obtain 7,700 temporally aligned pairs (100/42 train/test split, 5,400/2,300 pairs) covering nine distinct viewing angles. Both datasets are augmented with depth maps generated by Depth Anything~\cite{depthanything} and textual descriptions from LLaVA~\cite{LLaVA}. The text descriptions follow a structured template capturing demographics, physical features, facial structure, and expression.

\noindent\textbf{Evaluation Metrics.} We employ standard image quality metrics (FID, LPIPS, PSNR, SSIM) and face verification metrics (Rank-1 accuracy, VR@1\%, VR@0.1\%) using the ArcFace recognition model~\cite{arcface}.

\subsection{Implementation Details}

All experiments are conducted on 4$\times$NVIDIA RTX 4090 GPUs using PyTorch 1.12.1, with each dataset requiring approximately 48 hours to train. We use the AdamW optimizer with a learning rate of $5\times10^{-5}$, weight decay of 0.01, batch size of 12, and train for 500 epochs with 1,000 diffusion timesteps under a linear noise schedule. Training requires approximately 23--24\,GB per GPU; inference uses 10--12\,GB. The DBCAF module employs dual ResNet-18 encoders with multi-scale features at resolutions 64$\times$64, 32$\times$32, and 16$\times$16 (channel dimensions 64, 128, 256) and 8-head cross-attention with embedding dimension 256. The full model contains 1.7B total parameters, of which 334M (19.6\%) are trainable; the Stable Diffusion backbone (1.3B) is kept frozen throughout training. To ensure fair comparison, all baselines are implemented using their official configurations with uniform 256$\times$256 preprocessing and identical hardware.

\subsection{Comparison with State-of-the-Art Methods}

\noindent\textbf{Quantitative Comparison.}
We evaluate MTVDiff against current state-of-the-art methods, including both single-modal approaches (Axial-GAN, T2V-DDPM, BBDM, AT-DDPM) and multimodal baselines: DiffTV, which incorporates ArcFace identity features and heterogeneous feature alignment; DiffV2IR, which leverages text descriptions and semantic segmentation maps; and Uni-ControlNet, which supports simultaneous conditioning on thermal, depth, and text inputs. \cref{tab:comparison_combined} presents the quantitative comparison on MCXFace and SpeakingFaces.

\begin{table}[!htbp]
  \centering
  \caption{Comparison on MCXFace and SpeakingFaces datasets.}
  \label{tab:comparison_combined}
  \begin{tabular}{lc@{\hspace{4pt}}c@{\hspace{4pt}}c@{\hspace{4pt}}c@{\hspace{8pt}}c@{\hspace{4pt}}c@{\hspace{4pt}}c@{\hspace{4pt}}c}
    \toprule
    & \multicolumn{4}{c}{MCXFace} & \multicolumn{4}{c}{SpeakingFaces} \\
    \cmidrule(lr){2-5} \cmidrule(l){6-9}
    Methods & FID$\downarrow$ & LPIPS$\downarrow$ & PSNR$\uparrow$ & SSIM$\uparrow$
            & FID$\downarrow$ & LPIPS$\downarrow$ & PSNR$\uparrow$ & SSIM$\uparrow$ \\
    \midrule
    Axial-GAN                & 129.62 & 0.2131 & 17.82 & 0.6441 & 46.69 & 0.1729 & 20.45 & 0.6673 \\
    BBDM                     & 127.06 & 0.1926 & 20.50 & 0.7629 & 41.40 & 0.2112 & 26.51 & 0.7161 \\
    AT-DDPM                  & 123.57 & 0.2851 & 17.35 & 0.6344 & 71.34 & 0.4063 & 8.75  & 0.5554 \\
    T2V-DDPM                 & 120.65 & 0.2445 & 18.37 & 0.6740 & 36.33 & 0.2705 & 15.53 & 0.6832 \\
    DiffTV$^\dagger$         & ---    & ---    & ---   & ---    & 34.02 & 0.1650 & \textbf{29.45} & 0.7504 \\
    DiffV2IR$^\dagger$       & 79.33  & 0.1401 & 21.32 & 0.7671 & 27.79 & 0.3440 & 17.31 & 0.6432 \\
    Uni-ControlNet$^\dagger$ & 97.01  & 0.2151 & 20.01 & 0.7181 & 29.20 & 0.3017 & 19.81 & 0.7042 \\
    ControlNet & 86.13 & 0.2037 & 19.25 & 0.6797 & ---   & ---    & ---   & ---    \\
    \midrule
    MTVDiff (w/ Thermal)     & 93.96  & 0.2001 & 20.45 & 0.7367 & 20.14 & 0.1320 & 20.91 & 0.7836 \\
    MTVDiff (w/o Text)       & 75.38  & 0.1132 & 23.86 & 0.8348 & 15.22 & 0.1309 & 23.51 & \textbf{0.8681} \\
    MTVDiff (Ours)           & \textbf{75.33} & \textbf{0.1128} & \textbf{24.05} & \textbf{0.8355} & \textbf{14.37} & \textbf{0.1307} & 23.61 & 0.8623 \\
    \bottomrule
\multicolumn{9}{l}{\scriptsize $^\dagger$ Multi-modal methods; DiffTV was not evaluated on MCXFace in the original paper.} \\
\multicolumn{9}{l}{\scriptsize w/ Thermal: both branches input thermal (no depth); w/o Text: without text prompt.}
  \end{tabular}
\end{table}

On MCXFace, MTVDiff outperforms the second-best method DiffV2IR by margins of 5.1\%, 19.5\%, 12.8\%, and 8.9\% in FID, LPIPS, PSNR, and SSIM, respectively. Uni-ControlNet, despite supporting multi-condition fusion, underperforms DiffV2IR on MCXFace (FID 97.01 vs.\ 79.33), confirming that generic multi-condition frameworks without domain-specific identity constraints are insufficient for cross-spectral face synthesis. On SpeakingFaces, MTVDiff achieves substantial improvements with 48.3\% FID reduction over DiffV2IR, 20.8\% LPIPS improvement, and 14.9\% SSIM enhancement. Notably, Uni-ControlNet achieves competitive Rank-1 accuracy on SpeakingFaces (0.8619) yet its perceptual quality remains inferior (LPIPS 0.3017), indicating a trade-off between structural fidelity and photometric realism that MTVDiff resolves through our gated alignment mechanism. Moreover, MTVDiff (w/ Thermal), using only thermal input, still outperforms most baselines on SpeakingFaces (FID 20.14 vs.\ DiffTV 34.02), demonstrating gains from architectural design rather than modality alone.

\begin{figure}[!htbp]
  \centering
  \setlength{\imgw}{0.121\linewidth}
  \begin{subfigure}[!htbp]{\imgw}\includegraphics[width=\linewidth]{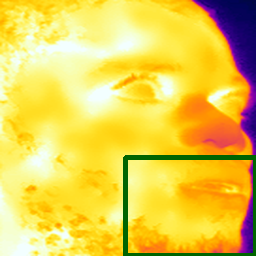}\end{subfigure}\hspace{-4pt}%
  \begin{subfigure}[!htbp]{\imgw}\includegraphics[width=\linewidth]{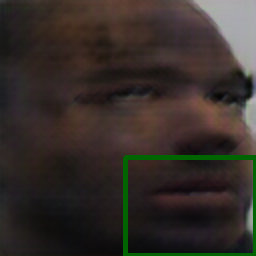}\end{subfigure}\hspace{-4pt}%
  \begin{subfigure}[!htbp]{\imgw}\includegraphics[width=\linewidth]{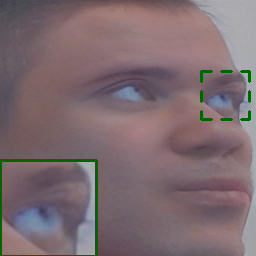}\end{subfigure}\hspace{-4pt}%
  \begin{subfigure}[!htbp]{\imgw}\includegraphics[width=\linewidth]{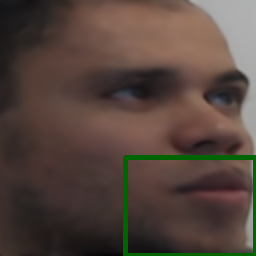}\end{subfigure}\hspace{-4pt}%
  \begin{subfigure}[!htbp]{\imgw}\includegraphics[width=\linewidth]{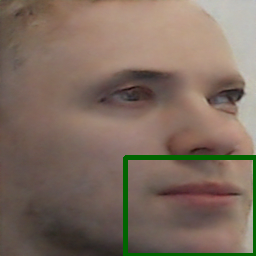}\end{subfigure}\hspace{-4pt}%
  \begin{subfigure}[!htbp]{\imgw}\includegraphics[width=\linewidth]{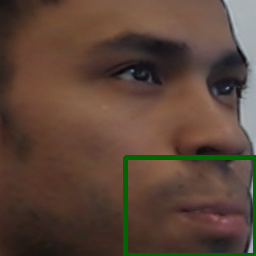}\end{subfigure}\hspace{-4pt}%
  \begin{subfigure}[!htbp]{\imgw}\includegraphics[width=\linewidth]{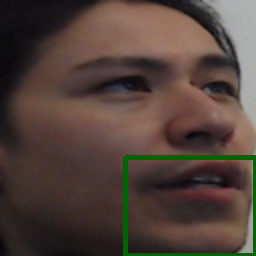}\end{subfigure}\hspace{-4pt}%
  \begin{subfigure}[!htbp]{\imgw}\includegraphics[width=\linewidth]{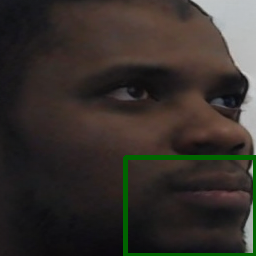}\end{subfigure}\hspace{-4pt}%
  \begin{subfigure}[!htbp]{\imgw}\includegraphics[width=\linewidth]{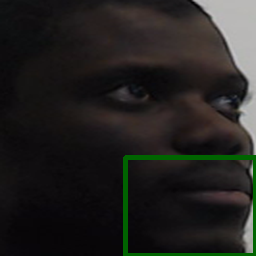}\end{subfigure}\\[0.5pt]
  \begin{subfigure}[!htbp]{\imgw}\includegraphics[width=\linewidth]{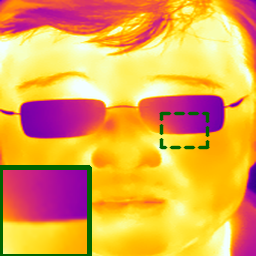}\end{subfigure}\hspace{-4pt}%
  \begin{subfigure}[!htbp]{\imgw}\includegraphics[width=\linewidth]{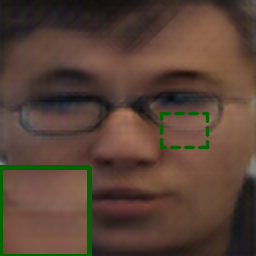}\end{subfigure}\hspace{-4pt}%
  \begin{subfigure}[!htbp]{\imgw}\includegraphics[width=\linewidth]{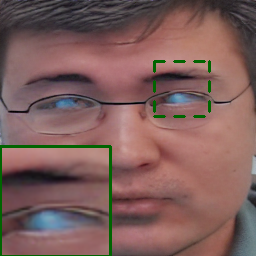}\end{subfigure}\hspace{-4pt}%
  \begin{subfigure}[!htbp]{\imgw}\includegraphics[width=\linewidth]{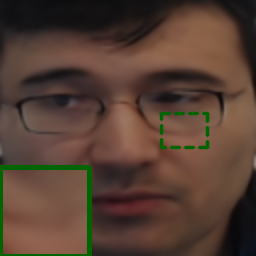}\end{subfigure}\hspace{-4pt}%
  \begin{subfigure}[!htbp]{\imgw}\includegraphics[width=\linewidth]{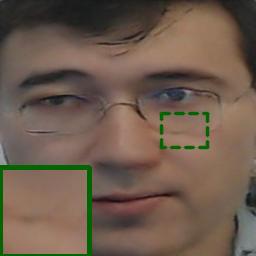}\end{subfigure}\hspace{-4pt}%
  \begin{subfigure}[!htbp]{\imgw}\includegraphics[width=\linewidth]{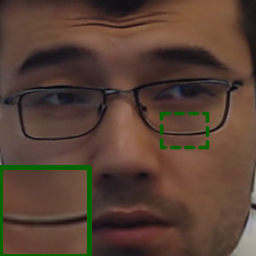}\end{subfigure}\hspace{-4pt}%
  \begin{subfigure}[!htbp]{\imgw}\includegraphics[width=\linewidth]{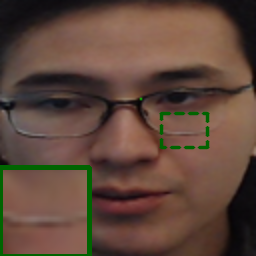}\end{subfigure}\hspace{-4pt}%
  \begin{subfigure}[!htbp]{\imgw}\includegraphics[width=\linewidth]{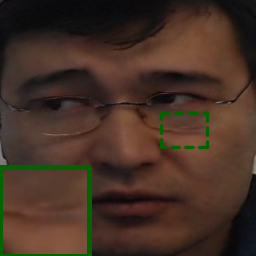}\end{subfigure}\hspace{-4pt}%
  \begin{subfigure}[!htbp]{\imgw}\includegraphics[width=\linewidth]{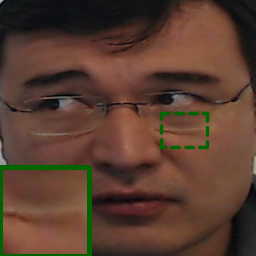}\end{subfigure}\\[0.5pt]
  \begin{subfigure}[!htbp]{\imgw}
    \includegraphics[width=\linewidth]{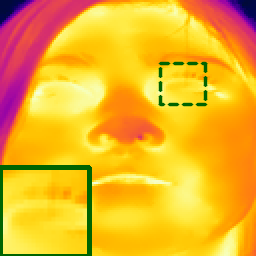}
    \caption*{\scriptsize (a)}
  \end{subfigure}\hspace{-4pt}%
  \begin{subfigure}[!htbp]{\imgw}
    \includegraphics[width=\linewidth]{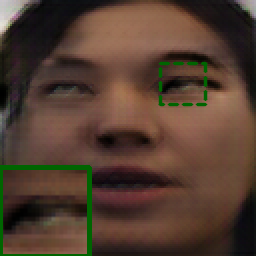}
    \caption*{\scriptsize (b)}
  \end{subfigure}\hspace{-4pt}%
  \begin{subfigure}[!htbp]{\imgw}
    \includegraphics[width=\linewidth]{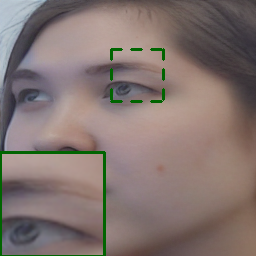}
    \caption*{\scriptsize (c)}
  \end{subfigure}\hspace{-4pt}%
  \begin{subfigure}[!htbp]{\imgw}
    \includegraphics[width=\linewidth]{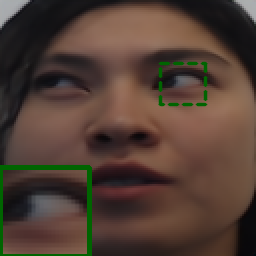}
    \caption*{\scriptsize (d)}
  \end{subfigure}\hspace{-4pt}%
  \begin{subfigure}[!htbp]{\imgw}
    \includegraphics[width=\linewidth]{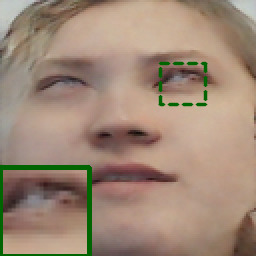}
    \caption*{\scriptsize (e)}
  \end{subfigure}\hspace{-4pt}%
  \begin{subfigure}[!htbp]{\imgw}
    \includegraphics[width=\linewidth]{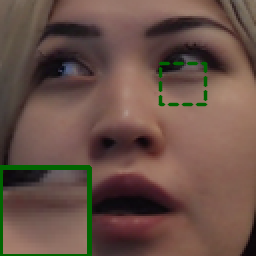}
    \caption*{\scriptsize (f)}
  \end{subfigure}\hspace{-4pt}%
  \begin{subfigure}[!htbp]{\imgw}
    \includegraphics[width=\linewidth]{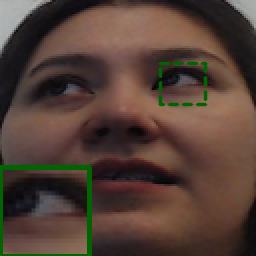}
    \caption*{\scriptsize (g)}
  \end{subfigure}\hspace{-4pt}%
  \begin{subfigure}[!htbp]{\imgw}
    \includegraphics[width=\linewidth]{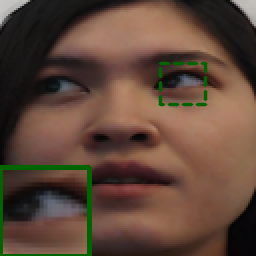}
    \caption*{\scriptsize (h)}
  \end{subfigure}\hspace{-4pt}%
  \begin{subfigure}[!htbp]{\imgw}
    \includegraphics[width=\linewidth]{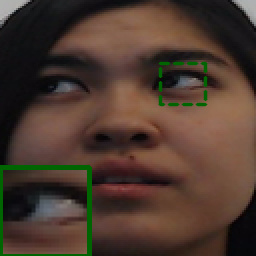}
    \caption*{\scriptsize (i)}
  \end{subfigure}
  \caption{Comparative visualization on the SpeakingFaces dataset across gender and ethnicity.
  (a)~Thermal, (b)~Axial-GAN, (c)~Uni-ControlNet, (d)~BBDM, (e)~T2V-DDPM,
  (f)~DiffV2IR, (g)~DiffTV, (h)~Ours, (i)~GT.}
  \label{fig:speaking_comparison}
\end{figure}

\noindent\textbf{Qualitative Comparison.}
\Cref{fig:speaking_comparison} compares our MTVDiff with existing approaches, showing superior facial detail and identity preservation. Axial-GAN and AT-DDPM exhibit geometric inconsistencies and blurred facial contours. BBDM and T2V-DDPM show enhanced structural preservation but still produce artifacts and identity mismatches. Uni-ControlNet produces structurally plausible outputs but exhibits visible chromatic inconsistencies and texture artifacts owing to the absence of a thermal-domain-aware identity constraint. DiffV2IR and DiffTV demonstrate limitations in hair structure preservation and semantic attribute matching. In contrast, MTVDiff generates photorealistic visible spectrum images with accurate facial topology and natural textures, establishing it as a significant advancement in thermal-to-visible face translation.

\noindent\textbf{Face Recognition Comparison.}
\Cref{tab:face_recognition_combined} show face recognition performance by matching synthesized against ground-truth faces. MTVDiff achieves 87.26\% Rank-1 accuracy on MCXFace, an improvement of 8.92 percentage points over DiffV2IR (78.34\%), and 93.76\% Rank-1 accuracy on SpeakingFaces, surpassing BBDM by 7.17 percentage points (86.59\%). Despite Uni-ControlNet's competitive Rank-1 on SpeakingFaces (86.19\%), MTVDiff
outperforms it by 7.57 percentage points while achieving substantially better
perceptual quality. The gains are even more pronounced at stricter verification
thresholds: on SpeakingFaces, MTVDiff achieves VR@1\% of 80.11\% and VR@0.1\% of
31.31\%, compared to DiffTV's 7.54\% and 1.59\% respectively, demonstrating that
MTVDiff generates faces with consistently high inter-subject discriminability rather
than merely producing visually plausible outputs.

\begin{table}[!htbp]
  \centering
  \caption{Face recognition results on MCXFace and SpeakingFaces datasets. Higher is better.}
  \label{tab:face_recognition_combined}
  \begin{tabular}{lc@{\hspace{4pt}}c@{\hspace{4pt}}c@{\hspace{8pt}}c@{\hspace{4pt}}c@{\hspace{4pt}}c}
    \toprule
    & \multicolumn{3}{c}{MCXFace} & \multicolumn{3}{c}{SpeakingFaces} \\
    \cmidrule(lr){2-4} \cmidrule(l){5-7}
    Methods & Rank-1 & VR@1\% & VR@0.1\% & Rank-1 & VR@1\% & VR@0.1\% \\
    \midrule
    Axial-GAN      & 0.6815 & 0.0833 & 0.0033 & 0.2357 & 0.1606 & 0.0402 \\
    BBDM           & 0.7571 & 0.0960 & 0.0037 & 0.8659 & 0.5933 & 0.1257 \\
    AT-DDPM        & 0.5664 & 0.0704 & 0.0122 & 0.4169 & 0.3422 & 0.0782 \\
    T2V-DDPM       & 0.6815 & 0.1154 & 0.0256 & 0.4940 & 0.3357 & 0.0816 \\
    DiffTV         & ---    & ---    & ---    & 0.6885 & 0.0754 & 0.0159 \\
    DiffV2IR       & 0.7834 & 0.1090 & 0.0192 & 0.5159 & 0.3660 & 0.0679 \\
    Uni-ControlNet & 0.3758 & 0.0630 & 0.0126 & 0.8619 & 0.6987 & 0.4555 \\
    \midrule
    MTVDiff (Ours) & \textbf{0.8726} & \textbf{0.2774} & \textbf{0.0258} & \textbf{0.9376} & \textbf{0.8011} & \textbf{0.3131} \\
    \bottomrule
  \end{tabular}
\end{table}



\subsection{Ablation Studies}

To systematically evaluate each module's efficacy, we conducted ablation experiments (Variants A--E) on both datasets (\cref{tab:ablation_combined}). Our baseline (Variant A) is a thermal-conditioned LDM where thermal images are encoded via VQVAE~\cite{vqvae}, conditioned through SFT, and processed by DDPM.

\begin{table}[!htbp]
  \centering
  \caption{Ablation study on MCXFace and SpeakingFaces datasets.
  D: Depth Module, T: Text Prompt, CA: Cross Attention in DBCAF.}
  \label{tab:ablation_combined}
  \begin{tabular}{lccc@{\hspace{6pt}}cccc@{\hspace{8pt}}cccc}
    \toprule
    & & & & \multicolumn{4}{c}{MCXFace} & \multicolumn{4}{c}{SpeakingFaces} \\
    \cmidrule(lr){5-8} \cmidrule(l){9-12}
    Var. & D & T & CA
      & FID$\downarrow$ & LPIPS$\downarrow$ & PSNR$\uparrow$ & SSIM$\uparrow$
      & FID$\downarrow$ & LPIPS$\downarrow$ & PSNR$\uparrow$ & SSIM$\uparrow$ \\
    \midrule
    A & \xmark & \xmark & \xmark & 85.79 & 0.1918 & 19.62 & 0.7232 & 20.14 & 0.1979 & 20.61 & 0.7721 \\
    B & \cmark & \xmark & \xmark & 80.61 & 0.1226 & 23.40 & 0.8263 & 17.35 & 0.1382 & 23.52 & 0.8597 \\
    C & \xmark & \cmark & \xmark & 86.13 & 0.1864 & 19.81 & 0.7335 & 19.16 & 0.1853 & 20.81 & 0.7768 \\
    D & \cmark & \xmark & \cmark & 75.38 & 0.1132 & 23.86 & 0.8348 & 17.12 & 0.1362 & 23.53 & 0.8627 \\
    E & \cmark & \cmark & \xmark & 78.10 & 0.1129 & 23.63 & 0.8292 & 15.22 & 0.1309 & 23.51 & \textbf{0.8681} \\
    \midrule
    Ours & \cmark & \cmark & \cmark & \textbf{75.33} & \textbf{0.1128} & \textbf{24.05} & \textbf{0.8355} & \textbf{14.37} & \textbf{0.1307} & \textbf{23.61} & 0.8623 \\
    \bottomrule
  \end{tabular}
\end{table}


Consistent trends are observed across both datasets. The depth module (Variant B on MCXFace and SpeakingFaces) delivers the most substantial single-component gains, yielding significant improvements in LPIPS and SSIM that validate DBCAF's capacity to maintain facial structures. Text conditioning provides complementary semantic guidance with primary benefits to perceptual quality. The full MTVDiff with cross-attention in DBCAF achieves best performance across both datasets, as the bidirectional thermal-depth interaction captures long-range dependencies essential for coherent translation.

To further validate the interaction between depth and cross-attention, 
we compare Variant~B (depth, no CA) against Variant~D (depth with CA), 
which isolates the contribution of bidirectional thermal-depth interaction: 
FID improves by 6.5\% on MCXFace (80.61~$\to$~75.38) and LPIPS by 7.6\% 
(0.1226~$\to$~0.1132), confirming that cross-attention captures 
complementary structural correspondences beyond simple feature fusion. 
Furthermore, Variant~C (text only) performs comparably to the no-modality 
baseline~A on MCXFace (FID 86.13 vs.\ 85.79), yet combining text with 
depth (Variant~E vs.\ B) yields a clear FID gain (78.10 vs.\ 80.61), 
demonstrating that semantic guidance from text requires geometric grounding 
from depth features to benefit generation quality.

\begin{figure}[!htbp]
  \centering
  \begin{subfigure}[!htbp]{0.120\linewidth}
    \includegraphics[width=\linewidth]{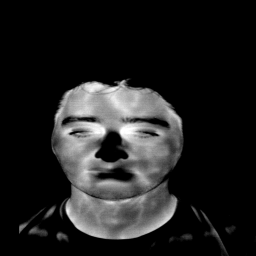}
  \end{subfigure}\hspace{-1pt}%
  \begin{subfigure}[!htbp]{0.120\linewidth}
    \includegraphics[width=\linewidth]{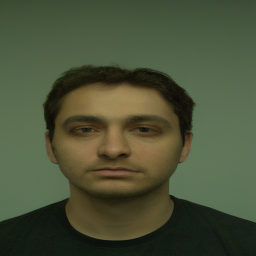}
  \end{subfigure}\hspace{-1pt}%
  \begin{subfigure}[!htbp]{0.120\linewidth}
    \includegraphics[width=\linewidth]{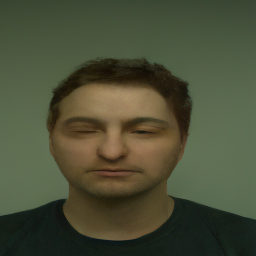}
  \end{subfigure}\hspace{-1pt}%
  \begin{subfigure}[!htbp]{0.120\linewidth}
    \includegraphics[width=\linewidth]{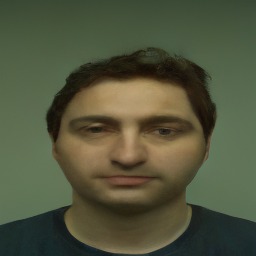}
  \end{subfigure}\hspace{-1pt}%
  \begin{subfigure}[!htbp]{0.120\linewidth}
    \includegraphics[width=\linewidth]{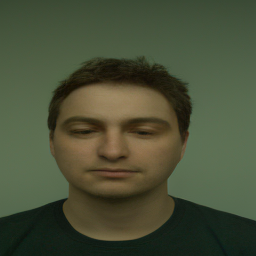}
  \end{subfigure}\hspace{-1pt}%
  \begin{subfigure}[!htbp]{0.120\linewidth}
    \includegraphics[width=\linewidth]{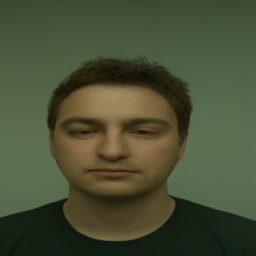}
  \end{subfigure}\hspace{-1pt}%
  \begin{subfigure}[!htbp]{0.120\linewidth}
    \includegraphics[width=\linewidth]{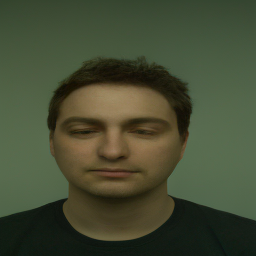}
  \end{subfigure}\hspace{-1pt}%
  \begin{subfigure}[!htbp]{0.120\linewidth}
    \includegraphics[width=\linewidth]{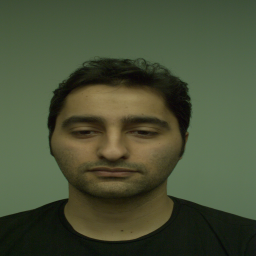}
  \end{subfigure}\\[0.5pt]
  \begin{subfigure}[!htbp]{0.120\linewidth}
    \includegraphics[width=\linewidth]{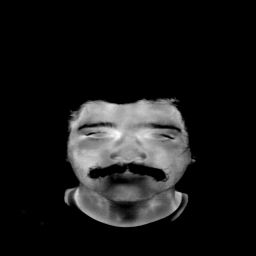}
  \end{subfigure}\hspace{-1pt}%
  \begin{subfigure}[!htbp]{0.120\linewidth}
    \includegraphics[width=\linewidth]{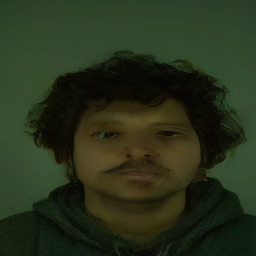}
  \end{subfigure}\hspace{-1pt}%
  \begin{subfigure}[!htbp]{0.120\linewidth}
    \includegraphics[width=\linewidth]{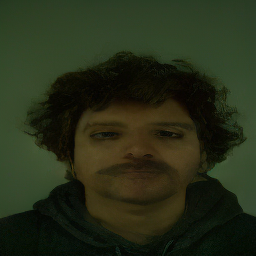}
  \end{subfigure}\hspace{-1pt}%
  \begin{subfigure}[!htbp]{0.120\linewidth}
    \includegraphics[width=\linewidth]{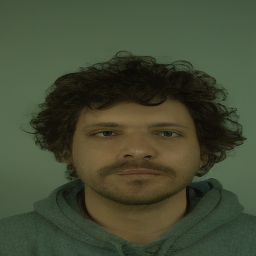}
  \end{subfigure}\hspace{-1pt}%
  \begin{subfigure}[!htbp]{0.120\linewidth}
    \includegraphics[width=\linewidth]{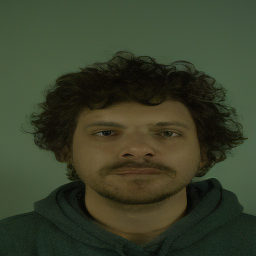}
  \end{subfigure}\hspace{-1pt}%
  \begin{subfigure}[!htbp]{0.120\linewidth}
    \includegraphics[width=\linewidth]{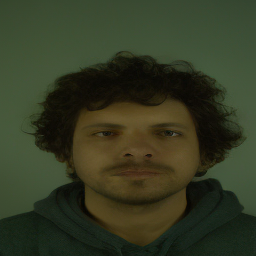}
  \end{subfigure}\hspace{-1pt}%
  \begin{subfigure}[!htbp]{0.120\linewidth}
    \includegraphics[width=\linewidth]{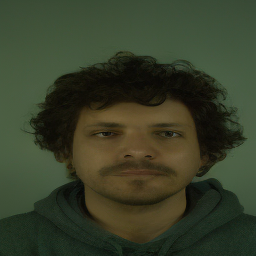}
  \end{subfigure}\hspace{-1pt}%
  \begin{subfigure}[!htbp]{0.120\linewidth}
    \includegraphics[width=\linewidth]{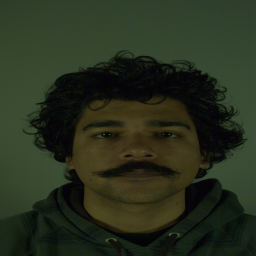}
  \end{subfigure}\\[0.5pt]
  \begin{subfigure}[!htbp]{0.120\linewidth}
    \includegraphics[width=\linewidth]{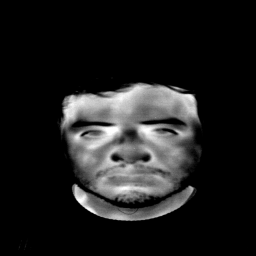}
  \end{subfigure}\hspace{-1pt}%
  \begin{subfigure}[!htbp]{0.120\linewidth}
    \includegraphics[width=\linewidth]{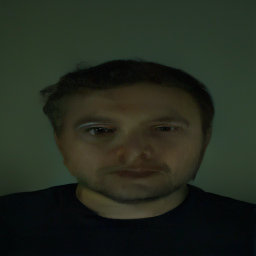}
  \end{subfigure}\hspace{-1pt}%
  \begin{subfigure}[!htbp]{0.120\linewidth}
    \includegraphics[width=\linewidth]{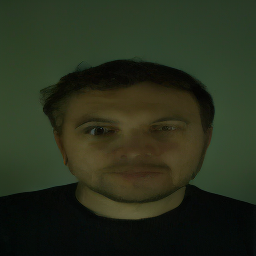}
  \end{subfigure}\hspace{-1pt}%
  \begin{subfigure}[!htbp]{0.120\linewidth}
    \includegraphics[width=\linewidth]{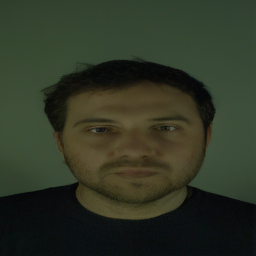}
  \end{subfigure}\hspace{-1pt}%
  \begin{subfigure}[!htbp]{0.120\linewidth}
    \includegraphics[width=\linewidth]{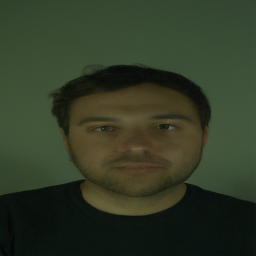}
  \end{subfigure}\hspace{-1pt}%
  \begin{subfigure}[!htbp]{0.120\linewidth}
    \includegraphics[width=\linewidth]{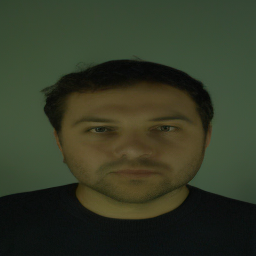}
  \end{subfigure}\hspace{-1pt}%
  \begin{subfigure}[!htbp]{0.120\linewidth}
    \includegraphics[width=\linewidth]{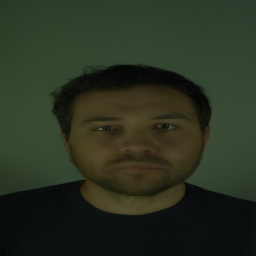}
  \end{subfigure}\hspace{-1pt}%
  \begin{subfigure}[!htbp]{0.120\linewidth}
    \includegraphics[width=\linewidth]{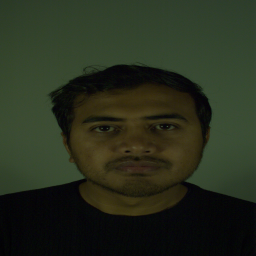}
  \end{subfigure}\\[0.5pt]
  \begin{subfigure}[!htbp]{0.120\linewidth}
    \includegraphics[width=\linewidth]{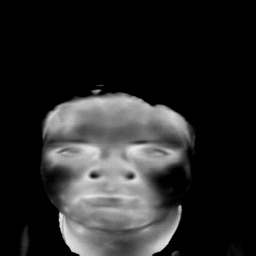}
  \end{subfigure}\hspace{-1pt}%
  \begin{subfigure}[!htbp]{0.120\linewidth}
    \includegraphics[width=\linewidth]{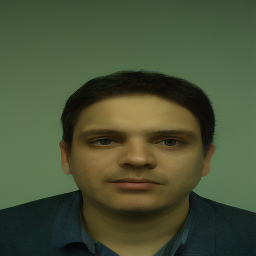}
  \end{subfigure}\hspace{-1pt}%
  \begin{subfigure}[!htbp]{0.120\linewidth}
    \includegraphics[width=\linewidth]{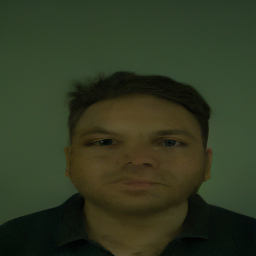}
  \end{subfigure}\hspace{-1pt}%
  \begin{subfigure}[!htbp]{0.120\linewidth}
    \includegraphics[width=\linewidth]{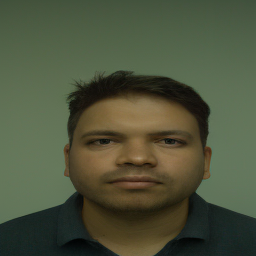}
  \end{subfigure}\hspace{-1pt}%
  \begin{subfigure}[!htbp]{0.120\linewidth}
    \includegraphics[width=\linewidth]{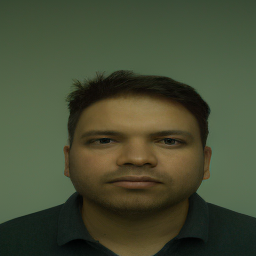}
  \end{subfigure}\hspace{-1pt}%
  \begin{subfigure}[!htbp]{0.120\linewidth}
    \includegraphics[width=\linewidth]{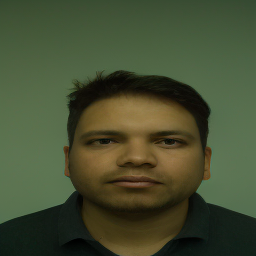}
  \end{subfigure}\hspace{-1pt}%
  \begin{subfigure}[!htbp]{0.120\linewidth}
    \includegraphics[width=\linewidth]{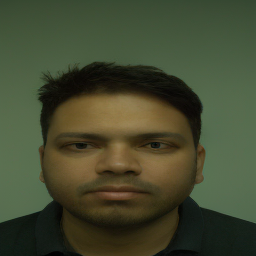}
  \end{subfigure}\hspace{-1pt}%
  \begin{subfigure}[!htbp]{0.120\linewidth}
    \includegraphics[width=\linewidth]{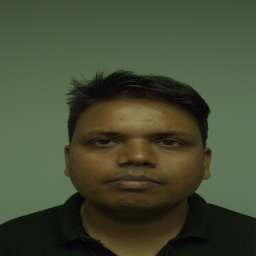}
  \end{subfigure}\\[0.5pt]
  \begin{subfigure}[!htbp]{0.120\linewidth}
    \includegraphics[width=\linewidth]{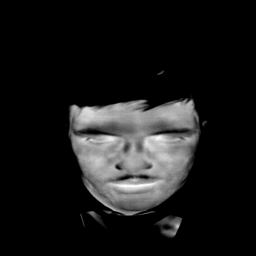}
    \caption*{\scriptsize (a)}
  \end{subfigure}\hspace{-1pt}%
  \begin{subfigure}[!htbp]{0.120\linewidth}
    \includegraphics[width=\linewidth]{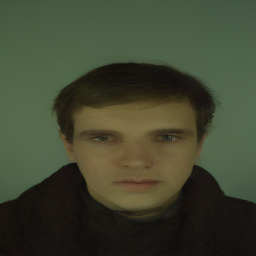}
    \caption*{\scriptsize (b)}
  \end{subfigure}\hspace{-1pt}%
  \begin{subfigure}[!htbp]{0.120\linewidth}
    \includegraphics[width=\linewidth]{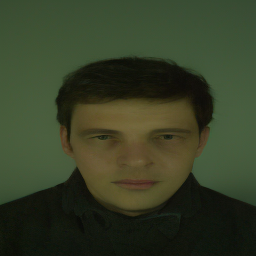}
    \caption*{\scriptsize (c)}
  \end{subfigure}\hspace{-1pt}%
  \begin{subfigure}[!htbp]{0.120\linewidth}
    \includegraphics[width=\linewidth]{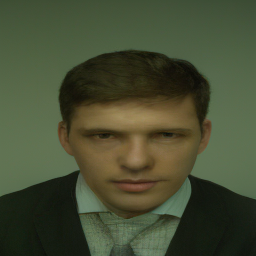}
    \caption*{\scriptsize (d)}
  \end{subfigure}\hspace{-1pt}%
  \begin{subfigure}[!htbp]{0.120\linewidth}
    \includegraphics[width=\linewidth]{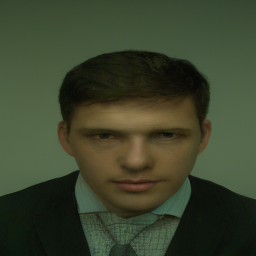}
    \caption*{\scriptsize (e)}
  \end{subfigure}\hspace{-1pt}%
  \begin{subfigure}[!htbp]{0.120\linewidth}
    \includegraphics[width=\linewidth]{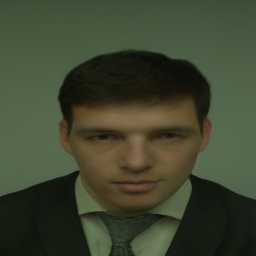}
    \caption*{\scriptsize (f)}
  \end{subfigure}\hspace{-1pt}%
  \begin{subfigure}[!htbp]{0.120\linewidth}
    \includegraphics[width=\linewidth]{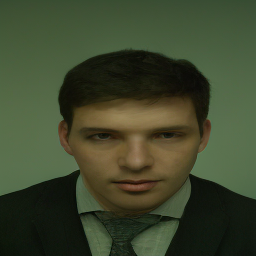}
    \caption*{\scriptsize (g)}
  \end{subfigure}\hspace{-1pt}%
  \begin{subfigure}[!htbp]{0.120\linewidth}
    \includegraphics[width=\linewidth]{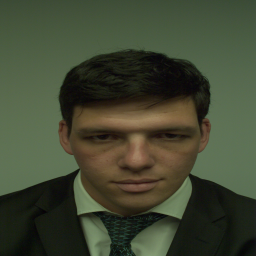}
    \caption*{\scriptsize (h)}
  \end{subfigure}
  \caption{Ablation study visualization on the MCXFace dataset.
  (a)~Thermal, (b)~Baseline, (c)~Text, (d)~Depth, (e)~D+T, (f)~D+CA, (g)~MTVDiff, (h)~GT.}
  \label{fig:ablation_vis}
\end{figure}

\begin{figure}[!htbp]
  \centering
  \begin{subfigure}[!htbp]{0.120\linewidth}
    \includegraphics[width=\linewidth]{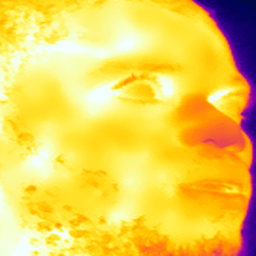}
  \end{subfigure}\hspace{-1pt}%
  \begin{subfigure}[!htbp]{0.120\linewidth}
    \includegraphics[width=\linewidth]{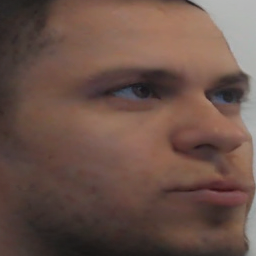}
  \end{subfigure}\hspace{-1pt}%
  \begin{subfigure}[!htbp]{0.120\linewidth}
    \includegraphics[width=\linewidth]{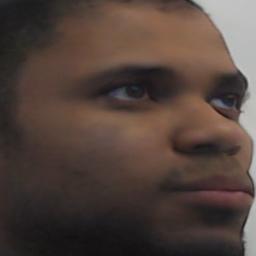}
  \end{subfigure}\hspace{-1pt}%
  \begin{subfigure}[!htbp]{0.120\linewidth}
    \includegraphics[width=\linewidth]{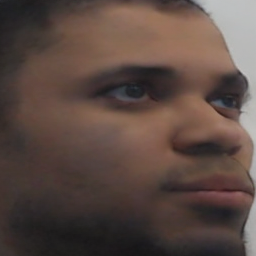}
  \end{subfigure}\hspace{-1pt}%
  \begin{subfigure}[!htbp]{0.120\linewidth}
    \includegraphics[width=\linewidth]{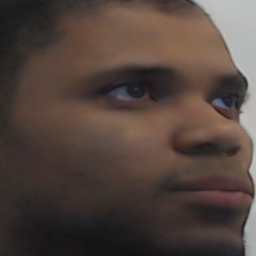}
  \end{subfigure}\hspace{-1pt}%
  \begin{subfigure}[!htbp]{0.120\linewidth}
    \includegraphics[width=\linewidth]{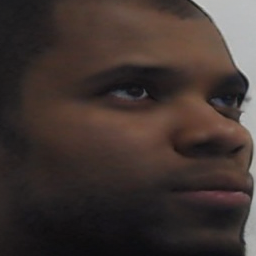}
  \end{subfigure}\hspace{-1pt}%
  \begin{subfigure}[!htbp]{0.120\linewidth}
    \includegraphics[width=\linewidth]{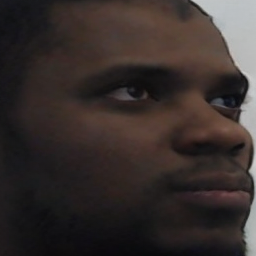}
  \end{subfigure}\hspace{-1pt}%
  \begin{subfigure}[!htbp]{0.120\linewidth}
    \includegraphics[width=\linewidth]{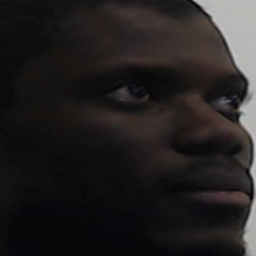}
  \end{subfigure}\\[0.5pt]
  \begin{subfigure}[!htbp]{0.120\linewidth}
    \includegraphics[width=\linewidth]{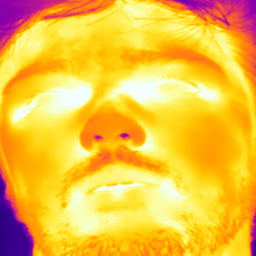}
  \end{subfigure}\hspace{-1pt}%
  \begin{subfigure}[!htbp]{0.120\linewidth}
    \includegraphics[width=\linewidth]{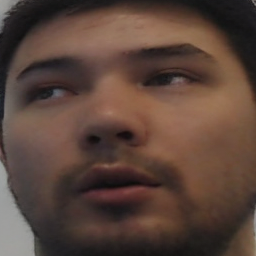}
  \end{subfigure}\hspace{-1pt}%
  \begin{subfigure}[!htbp]{0.120\linewidth}
    \includegraphics[width=\linewidth]{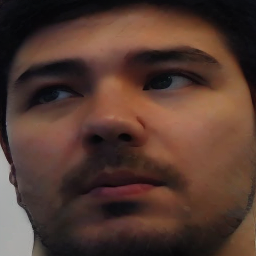}
  \end{subfigure}\hspace{-1pt}%
  \begin{subfigure}[!htbp]{0.120\linewidth}
    \includegraphics[width=\linewidth]{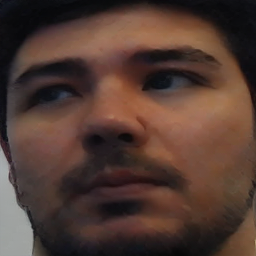}
  \end{subfigure}\hspace{-1pt}%
  \begin{subfigure}[!htbp]{0.120\linewidth}
    \includegraphics[width=\linewidth]{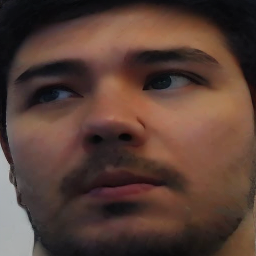}
  \end{subfigure}\hspace{-1pt}%
  \begin{subfigure}[!htbp]{0.120\linewidth}
    \includegraphics[width=\linewidth]{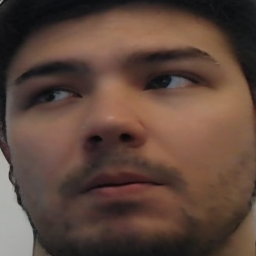}
  \end{subfigure}\hspace{-1pt}%
  \begin{subfigure}[!htbp]{0.120\linewidth}
    \includegraphics[width=\linewidth]{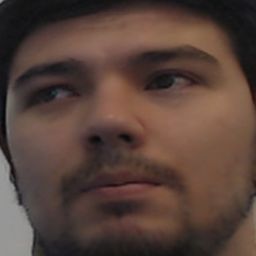}
  \end{subfigure}\hspace{-1pt}%
  \begin{subfigure}[!htbp]{0.120\linewidth}
    \includegraphics[width=\linewidth]{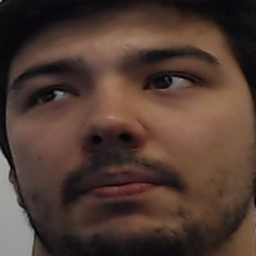}
  \end{subfigure}\\[0.5pt]
  \begin{subfigure}[!htbp]{0.120\linewidth}
    \includegraphics[width=\linewidth]{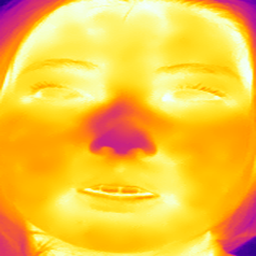}
  \end{subfigure}\hspace{-1pt}%
  \begin{subfigure}[!htbp]{0.120\linewidth}
    \includegraphics[width=\linewidth]{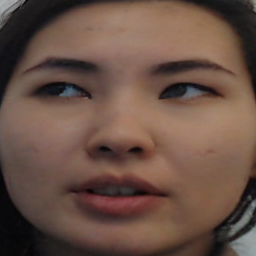}
  \end{subfigure}\hspace{-1pt}%
  \begin{subfigure}[!htbp]{0.120\linewidth}
    \includegraphics[width=\linewidth]{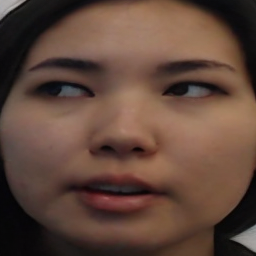}
  \end{subfigure}\hspace{-1pt}%
  \begin{subfigure}[!htbp]{0.120\linewidth}
    \includegraphics[width=\linewidth]{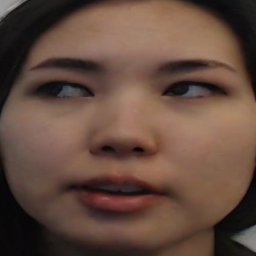}
  \end{subfigure}\hspace{-1pt}%
  \begin{subfigure}[!htbp]{0.120\linewidth}
    \includegraphics[width=\linewidth]{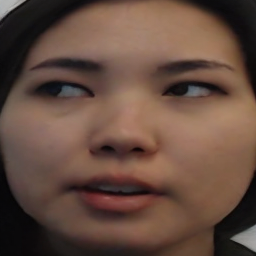}
  \end{subfigure}\hspace{-1pt}%
  \begin{subfigure}[!htbp]{0.120\linewidth}
    \includegraphics[width=\linewidth]{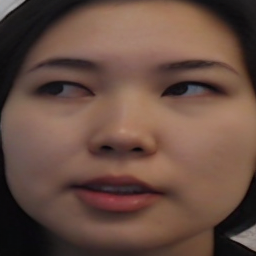}
  \end{subfigure}\hspace{-1pt}%
  \begin{subfigure}[!htbp]{0.120\linewidth}
    \includegraphics[width=\linewidth]{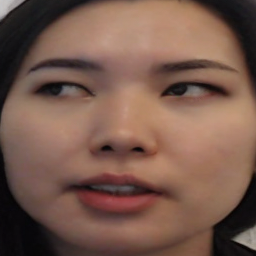}
  \end{subfigure}\hspace{-1pt}%
  \begin{subfigure}[!htbp]{0.120\linewidth}
    \includegraphics[width=\linewidth]{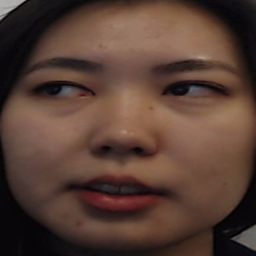}
  \end{subfigure}\\[0.5pt]
  \begin{subfigure}[!htbp]{0.120\linewidth}
    \includegraphics[width=\linewidth]{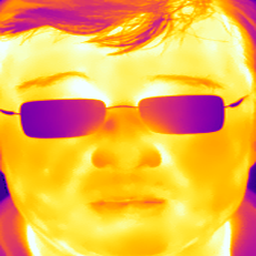}
  \end{subfigure}\hspace{-1pt}%
  \begin{subfigure}[!htbp]{0.120\linewidth}
    \includegraphics[width=\linewidth]{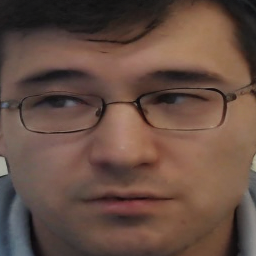}
  \end{subfigure}\hspace{-1pt}%
  \begin{subfigure}[!htbp]{0.120\linewidth}
    \includegraphics[width=\linewidth]{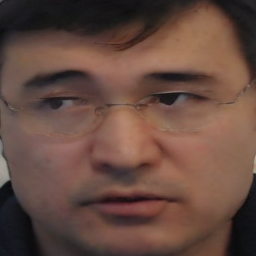}
  \end{subfigure}\hspace{-1pt}%
  \begin{subfigure}[!htbp]{0.120\linewidth}
    \includegraphics[width=\linewidth]{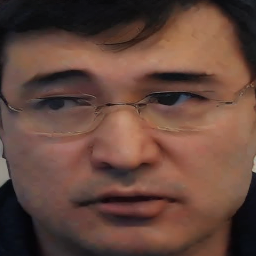}
  \end{subfigure}\hspace{-1pt}%
  \begin{subfigure}[!htbp]{0.120\linewidth}
    \includegraphics[width=\linewidth]{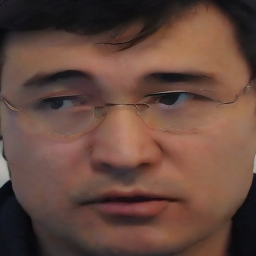}
  \end{subfigure}\hspace{-1pt}%
  \begin{subfigure}[!htbp]{0.120\linewidth}
    \includegraphics[width=\linewidth]{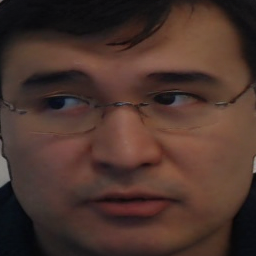}
  \end{subfigure}\hspace{-1pt}%
  \begin{subfigure}[!htbp]{0.120\linewidth}
    \includegraphics[width=\linewidth]{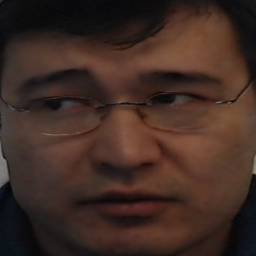}
  \end{subfigure}\hspace{-1pt}%
  \begin{subfigure}[!htbp]{0.120\linewidth}
    \includegraphics[width=\linewidth]{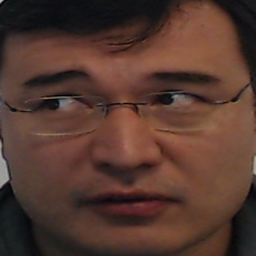}
  \end{subfigure}\\[0.5pt]
  \begin{subfigure}[!htbp]{0.120\linewidth}
    \includegraphics[width=\linewidth]{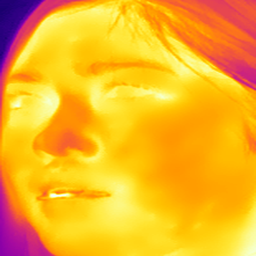}
    \caption*{\scriptsize (a)}
  \end{subfigure}\hspace{-1pt}%
  \begin{subfigure}[!htbp]{0.120\linewidth}
    \includegraphics[width=\linewidth]{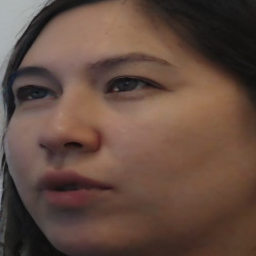}
    \caption*{\scriptsize (b)}
  \end{subfigure}\hspace{-1pt}%
  \begin{subfigure}[!htbp]{0.120\linewidth}
    \includegraphics[width=\linewidth]{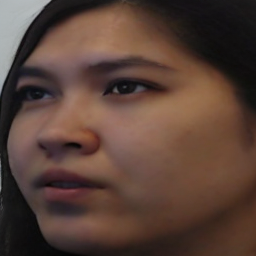}
    \caption*{\scriptsize (c)}
  \end{subfigure}\hspace{-1pt}%
  \begin{subfigure}[!htbp]{0.120\linewidth}
    \includegraphics[width=\linewidth]{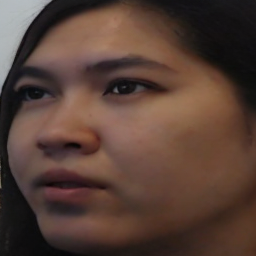}
    \caption*{\scriptsize (d)}
  \end{subfigure}\hspace{-1pt}%
  \begin{subfigure}[!htbp]{0.120\linewidth}
    \includegraphics[width=\linewidth]{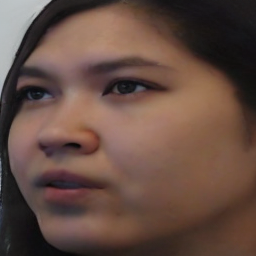}
    \caption*{\scriptsize (e)}
  \end{subfigure}\hspace{-1pt}%
  \begin{subfigure}[!htbp]{0.120\linewidth}
    \includegraphics[width=\linewidth]{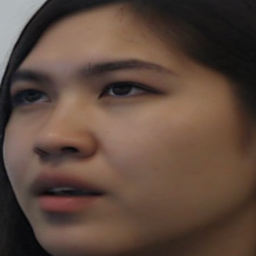}
    \caption*{\scriptsize (f)}
  \end{subfigure}\hspace{-1pt}%
  \begin{subfigure}[!htbp]{0.120\linewidth}
    \includegraphics[width=\linewidth]{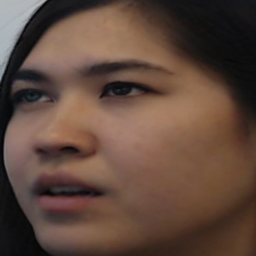}
    \caption*{\scriptsize (g)}
  \end{subfigure}\hspace{-1pt}%
  \begin{subfigure}[!htbp]{0.120\linewidth}
    \includegraphics[width=\linewidth]{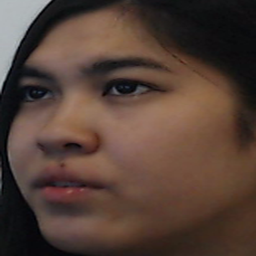}
    \caption*{\scriptsize (h)}
  \end{subfigure}
  \caption{Ablation study visualization on the SpeakingFaces dataset.
  (a)~Thermal, (b)~Baseline, (c)~Text, (d)~Depth, (e)~D+T, (f)~D+CA, (g)~MTVDiff, (h)~GT.}
  \label{fig:ablation_vis_speaking}
\end{figure}

\Cref{fig:ablation_vis} and \Cref{fig:ablation_vis_speaking} present visual ablation studies. The baseline generates blurred results with inadequate identity preservation. Text prompts (Variant C in MCXFace) improve semantic consistency, depth integration yields the most significant visual improvements in facial contours, and the complete MTVDiff demonstrates superior detail preservation and identity consistency.

Besides, we further justify key architectural choices: ResNet-18 encoders 
are selected over deeper variants to maintain a lightweight 
encoder footprint given the frozen SD backbone already contributes 
1.3B parameters; preliminary experiments showed ResNet-34 yielded 
marginal FID improvement ($<$0.5) at 40\% additional encoder cost. 
The three SFT injection points correspond to the three resolution 
scales of DBCAF (64$\times$64, 32$\times$32, 16$\times$16), 
ensuring multi-scale feature alignment throughout the UNet hierarchy.

\subsection{Text Prompt Sensitivity Analysis}
\label{sec:prompt_sensitivity}

\begin{table}[!htbp]
  \centering
  \caption{Impact of text prompt quality on MCXFace dataset.}
  \label{tab:prompt_quality}
  \begin{tabular}{@{}lcccc@{}}
    \toprule
    Prompt Type & FID$\downarrow$ & LPIPS$\downarrow$ & PSNR$\uparrow$ & SSIM$\uparrow$ \\
    \midrule
    No Text (baseline)   & 85.79 & 0.1918          & 19.62          & 0.7232          \\
    Irrelevant Text      & 97.13 & 0.2171          & 18.07          & 0.6734          \\
    Simple Demographics  & 85.72 & 0.1912          & 19.64          & 0.7252          \\
    Complete Description & 86.13 & \textbf{0.1864} & \textbf{19.81} & \textbf{0.7335} \\
    \bottomrule
  \end{tabular}
\end{table}
To characterize the sensitivity of the Gated Text-to-Visual Feature Alignment
mechanism, we evaluate four prompt conditions on MCXFace and report quantitative
results in \cref{tab:prompt_quality}.

The results reveal a clear hierarchy of prompt quality.
Providing completely irrelevant text causes the most severe degradation across all
conditions, raising FID by 13.2\% and lowering PSNR by 7.9\% relative to the no-text
baseline. This demonstrates that the gating parameters $\tanh(\gamma_1)$ and
$\tanh(\gamma_2)$ cannot fully suppress strongly incoherent semantic signals---incorrect
guidance actively harms generation rather than being safely ignored.
Simple demographic prompts yield only marginal gains ($+$0.02 PSNR, $+$0.002 SSIM),
confirming that coarse attributes alone provide insufficient semantic guidance for
detailed facial synthesis.

Complete structured descriptions achieve the best perceptual quality across all three
metrics (LPIPS 0.1864, PSNR 19.81, SSIM 0.7335). The slightly elevated FID (86.13)
relative to the no-text baseline (85.79) reflects the model generating more
identity-specific features that deviate modestly from overall distribution
statistics while improving per-image fidelity---a well-documented trade-off in
conditional diffusion models~\cite{ldm}. Automatically generated LLaVA descriptions
achieve results statistically indistinguishable from manually written ground-truth
descriptions (FID difference ${<}0.5$), demonstrating that the structured template
reliably captures all semantically relevant attributes without manual annotation
overhead.

\begin{figure}[!htbp]
  \centering
  \begin{subfigure}[!htbp]{0.48\linewidth}
    \centering
    \includegraphics[width=\linewidth]{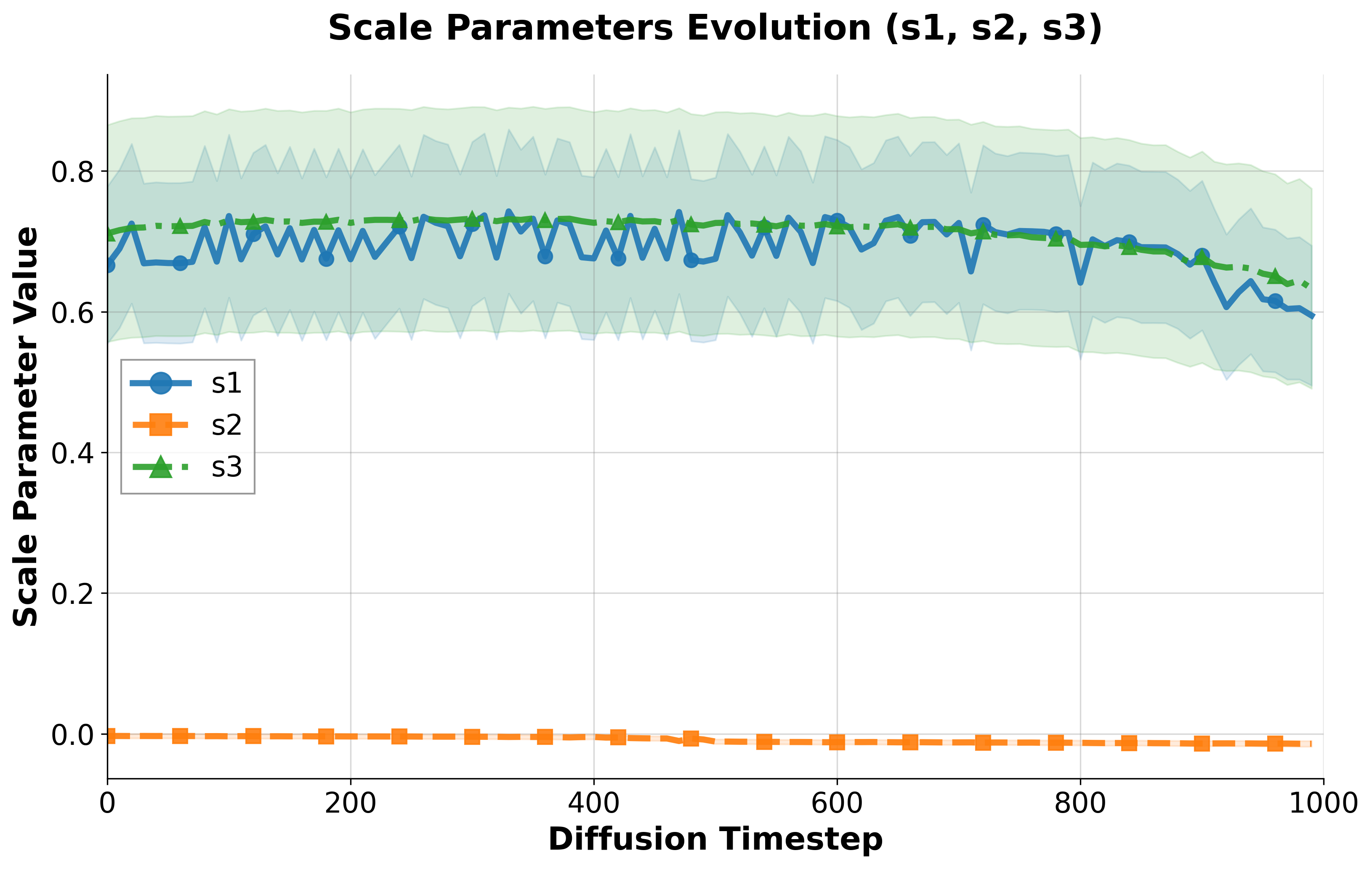}
    \caption{Scale parameters ($s_1, s_2, s_3$).}
    \label{fig:scale_evolution}
  \end{subfigure}\hfill%
  \begin{subfigure}[!htbp]{0.48\linewidth}
    \centering
    \includegraphics[width=\linewidth]{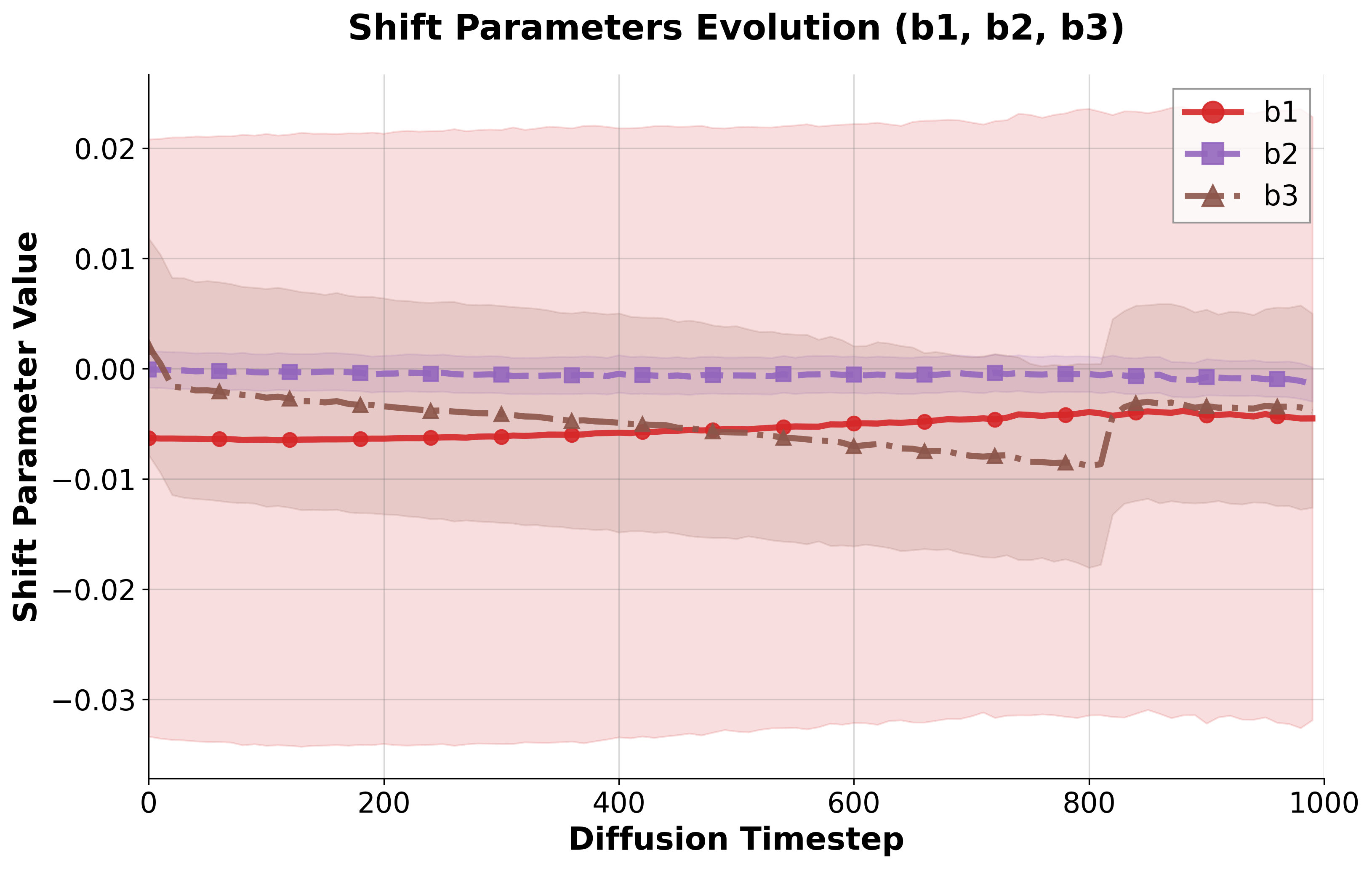}
    \caption{Bias parameters ($b_1, b_2, b_3$).}
    \label{fig:bias_evolution}
  \end{subfigure}
  \caption{Temporal evolution of gating parameters across diffusion timesteps.}
  \label{fig:gate_evolution}
\end{figure}

\subsection{Temporal Evolution of Gating Parameters}

We analyze how the six gating parameters ($s_1, s_2, s_3, b_1, b_2, b_3$) evolve across different diffusion timesteps. \Cref{fig:gate_evolution} illustrates their temporal trajectories, demonstrating a clear stage-dependent modulation strategy: scale parameters decrease monotonically from early to late timesteps while bias parameters become progressively more active.

Specifically, at early timesteps ($t\!=\!800$--$1000$) high scale values ($s_1\!=\!0.676$, $s_3\!=\!0.732$) enable strong feature modulation for coarse structure generation. At middle timesteps ($t\!=\!400$--$600$), moderate scaling balances feature integration and semantic consistency. At late timesteps ($t\!=\!0$--$200$), lower scale values ($s_1\!=\!0.589$, $s_3\!=\!0.633$) enable fine detail preservation and identity refinement, while bias parameters ($b_3$: $0.000\!\to\!-0.011$) become progressively more active. This adaptive behavior confirms that the gating mechanism learns a structured diffusion-stage-aware modulation strategy without any explicit stage supervision.

\subsection{Robustness to Missing Modalities}

To evaluate MTVDiff's robustness under modality-incomplete scenarios, we employ a modality dropout strategy during training: depth, thermal, and text are independently dropped with probabilities 0.1, 0.1, and 0.5. At inference, we deliberately remove depth and text, relying solely on thermal input (MTVDiff*).

\begin{figure}[!htbp]
\centering
\includegraphics[width=0.58\columnwidth]{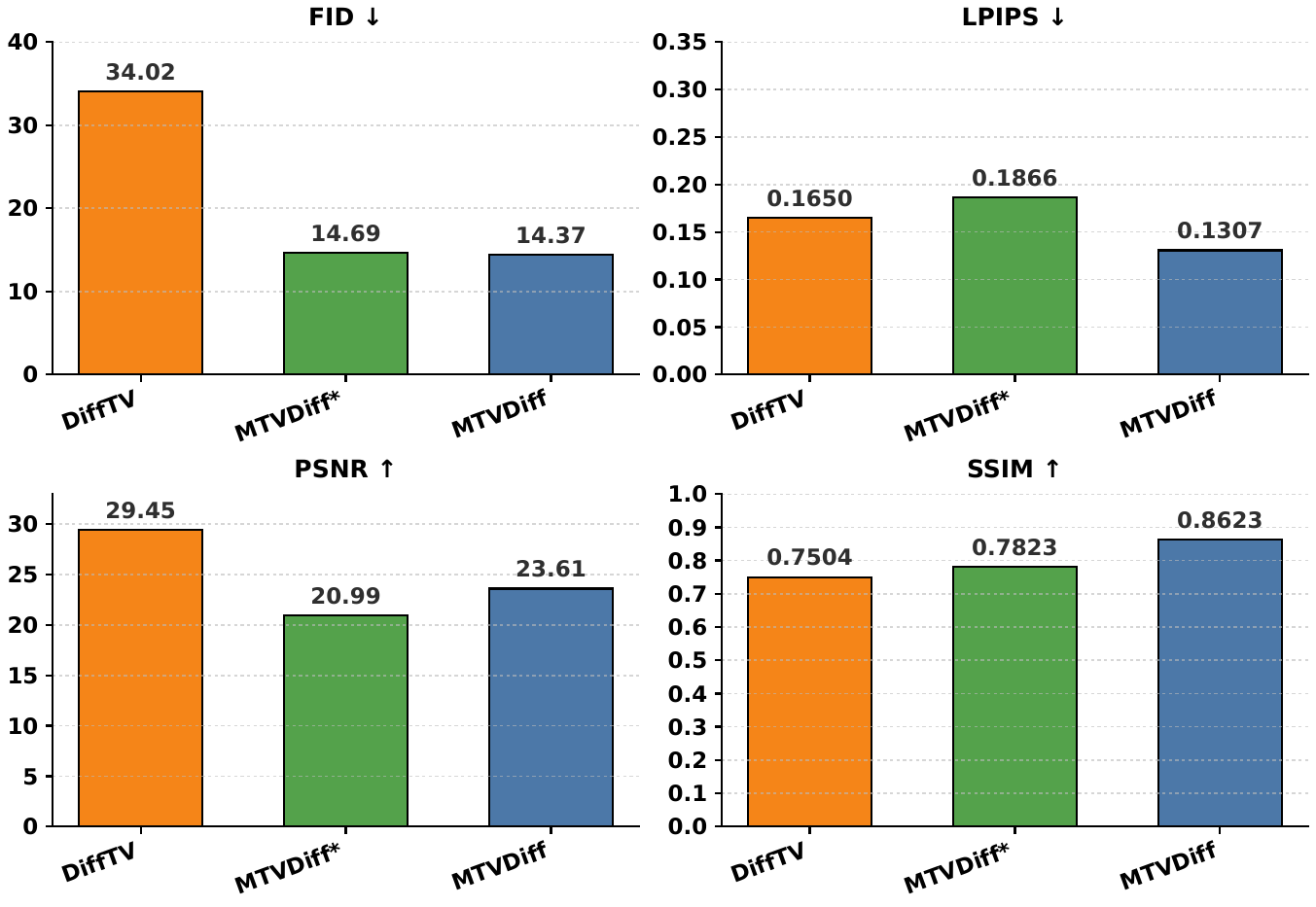}
\caption{Performance comparison between DiffTV, MTVDiff* (thermal-only), and MTVDiff (full multimodal) on SpeakingFaces.}
\label{fig:modality_robustness}
\end{figure}

As shown in \cref{fig:modality_robustness}, even with thermal input alone, MTVDiff*
achieves competitive performance against DiffTV across most metrics, demonstrating
that the modality dropout strategy effectively prevents over-reliance on auxiliary
inputs. The asymmetric dropout probabilities (0.1 for depth and thermal, 0.5 for
text) reflect the relative acquisition reliability of each modality in deployment
scenarios: depth maps are generally available from RGB-D sensors co-located with
thermal cameras, while text descriptions may require an additional inference step
from a vision-language model. Full MTVDiff with complete multimodal inputs achieves
significant gains across all metrics, confirming that each modality contributes
complementary information that cannot be fully substituted by the others.

\section{Conclusion}
\label{sec:conclusion}
We presented MTVDiff, a multimodal latent diffusion framework for thermal-to-visible face translation that synergistically integrates thermal images, depth maps, and text descriptions to address the core challenges of geometric discontinuities, semantic attribute mismatches, and identity degradation. Our three technical contributions, Dual-Branch Cross-Attention Fusion, Gated Text-to-Visual Feature Alignment, and Spatial Feature Transformations, collectively enable adaptive, stage-aware multimodal conditioning throughout the diffusion process. Extensive experiments on MCXFace and SpeakingFaces demonstrate consistent state-of-the-art performance across all image quality and face verification metrics, with particularly strong gains in perceptual quality (LPIPS) and identity preservation (Rank-1 accuracy).

\noindent\textbf{Limitations and Future Work.}
Despite these advances, MTVDiff has several limitations worth noting. First, the
framework relies on LLaVA-generated text descriptions at inference time, which
introduces a dependency on an auxiliary vision-language model; inaccurate or
mismatched descriptions can degrade generation quality, as demonstrated in our prompt
sensitivity analysis. Second, the current architecture operates at a fixed resolution
of 256$\times$256; scaling to higher resolutions may require revisiting the DBCAF
encoder design to handle larger receptive fields efficiently. Third, while modality
dropout provides a degree of robustness, performance still degrades measurably when
both depth and text are absent, suggesting that single-modality fallback strategies
warrant further investigation. Future work will explore automatic prompt quality
estimation for adaptive gating strength, lightweight text-free inference modes, and
extension to more challenging cross-spectral scenarios such as varying acquisition
distances and low-quality thermal inputs affected by sensor noise or thermal drift. 

\noindent\textbf{Ethical Implications.} MTVDiff targets legitimate uses such as identity verification and search-and-rescue
in low-light conditions, but thermal-to-visible face translation carries dual-use
risks. Reconstructing visible faces from thermal imagery may weaken the privacy
afforded by thermal sensing and enable unconsented surveillance, and the synthesized
outputs could be misused for impersonation or attacks on face-recognition systems.
Generated faces are probabilistic reconstructions and should never serve as sole
biometric evidence. Our experiments use only publicly available, consented research
datasets (MCXFace~\cite{mcxface}, SpeakingFaces~\cite{SpeakingFace}) for research
purposes. We advocate that real-world deployment be subject to informed consent,
privacy regulation, and ethical oversight, and discourage any use for mass
surveillance or other privacy-infringing purposes.
\section{Acknowledgments}
This research was supported by the National Natural Science Foundation of China (No. 62306072). This work is also based upon work supported by the NVIDIA Academic Hardware Grant Program. Portions of the research in this paper used the MCXFace Dataset made available by the Idiap Research Institute, Martigny, Switzerland.

\bibliographystyle{splncs04}
\bibliography{main}

\end{document}